\documentclass[letterpaper, 10 pt, conference]{ieeeconf}  

\IEEEoverridecommandlockouts                              

\usepackage{amssymb}  
\usepackage{graphics} 
\usepackage{graphicx}
\usepackage[tight,footnotesize]{subfigure}
\long\def\invis#1{}
\usepackage{cite}
\usepackage{url}
\usepackage{hyphenat}
\usepackage{amsmath,systeme}
\usepackage{xcolor}
\usepackage[linesnumbered,ruled]{algorithm2e}
\usepackage{siunitx}
\usepackage{enumerate}
\usepackage{mathtools}

\usepackage[utf8]{inputenc}
\usepackage{ wasysym }
\usepackage{pgfplots}
\usepackage{tikz}
\usepackage{pgfplotstable}
\usepgfplotslibrary{colormaps}
\pgfplotsset{colormap/hot}
\pgfplotsset{width=9cm,compat=1.9}
\pgfplotsset{layers/my layer set/.define layer set={background,main,foreground}{ },set layers=my layer set,}
\usepackage{xcolor}
\usepackage{tabularx}
\usepackage{multirow}
\usepackage{graphicx} 

\newcommand\bmat{\begin{bmatrix}}
\newcommand\emat{\end{bmatrix}}
\usepackage{textcomp}
\usepackage{svg}
\usepackage{algpseudocode}

\usepackage{fancyhdr}
\fancypagestyle{withfooter}{
  
  \fancyfoot[C]{\footnotesize Accepted to the IEEE IROS workshop on Autonomous Robotic Systems in Aquaculture: Research Challenges and Industry Needs}
}

\def\floatsep{4.0pt}
\def\textfloatsep{0.0pt}
\def\dblfloatsep{4.0pt}
\def\dbltextfloatsep{4.0pt}

\usepackage{fp}
\FPset{\pb}{0}
\newcommand{\pagebudget}[1]{}

\usepackage{xcolor}

\newif\ifmarkchange

\definecolor{changed}{rgb}{0.0,0.0,0.0}
\ifmarkchange
\definecolor{changed}{rgb}{0.91, 0.05, 0.05}
\fi

\title{\LARGE \bf Framework for Robust Localization of UUVs and Mapping of Net Pens

}

\author{David Botta$^1$, Luca Ebner$^2$, Andrej Studer$^2$, Victor Reijgwart$^1$, Roland Siegwart$^1$ and Eleni Kelasidi$^{1,3}$  
\thanks{$^1$ David Botta, Victor Reijgwart and Roland Siegwart are with the Department of Mechanical and Process Engineering, ETH {\tt\small bottad@student.ethz.ch, victor.reijgwart@mavt.ethz.ch, rolandsi@ethz.ch}}
\thanks{$^2$ Luca Ebner and Andrej Studer are with Tethys Robotics {\tt\small luca@tethys-robotics.ch, andrej@tethys-robotics.ch}}
\thanks{$^3$ Eleni Kelasidi is with the Department of Mechanical and Process Engineering, ETH and with the Aquaculture Robotics and Automation Group, SINTEF Ocean {\tt\small eleni.kelasidi@sintef.no}}
\thanks{This work was supported by the Research Council of Norway (ResiFarm: NO-327292, CHANGE: N313737).}}

\begin{document}

\maketitle
\thispagestyle{withfooter}
\pagestyle{withfooter}

\begin{abstract}
This paper presents a general framework integrating vision and acoustic sensor data to enhance localization and mapping in highly dynamic and complex underwater environments, with a particular focus on fish farming. The proposed pipeline is suited to obtain both the net-relative pose estimates of an Unmanned Underwater Vehicle (UUV) and the depth map of the net pen purely based on vision data. Furthermore, this paper presents a method to estimate the global pose of an UUV fusing the net-relative pose estimates with acoustic data. The pipeline proposed in this paper showcases results on datasets obtained from industrial-scale fish farms and successfully demonstrates that the vision-based TRU-Depth model, when provided with sparse depth priors from the FFT method and combined with the Wavemap method, can estimate both net-relative and global position of the UUV in real time and generate detailed 3D maps suitable for autonomous navigation and inspection purposes.
\end{abstract}
\section{Introduction}
\label{sec:intro}
The aquaculture industry has seen rapid growth over the last decades.
Fish farming, in particular, has emerged as a vital source of the global seafood supply \cite{FAO2018}. However, the rapid growth of this industry presents new challenges, especially in terms of ensuring efficient, safe, and sustainable operations \cite{fore2018precision}. Fish farming operations often involve a significant amount of manual labor, which can be both physically demanding and dangerous. Tasks such as net inspection, maintenance, and repairs expose workers to hazardous underwater conditions, including rough seas, low visibility, and the presence of potentially harmful marine life. Addressing some of these problems, the adaptation of robotic systems in aquaculture has also grown significantly in recent years \cite{kelasidi2023robotics}.

Current robotic solutions often involve the use of manually operated UUVs, such as Remotely Operated Vehicles (ROVs) for inspection and intervention operations in fish farms, which are expensive to deploy as they can only be operated by highly trained ROV pilots~\cite{fore2018precision,kelasidi2023robotics}. As the number of fish farms increases, and with the trend toward deploying these farms in increasingly remote locations \cite{Bjelland2015}, the automation of such tasks becomes crucial for enhancing operational efficiency \cite{schellewald2021vision}. Autonomous UUVs offer a promising solution to these challenges, thus reducing weather and manual labor-dependent risks and allowing for more efficient and sustainable operations~\cite{kelasidi2023robotics}. However, the effective deployment of autonomous UUVs in fish farms requires robust and precise localization and mapping methods within the net pens, which still remains an open research problem.

Traditional UUV navigation systems rely heavily on acoustic sensors, such as echo-sounders, Ultra-short baseline (USBL) acoustic positioning systems, and Doppler Velocity Loggers (DVL) \cite{Fossen2011,kelasidi2023robotics}. While effective in many underwater scenarios, these sensors face significant limitations in fish farms. The permeable nature of fish nets can result in weak or distorted acoustic reflections, leading to weak target signal strengths \cite{Amundsen2022}. Additionally, the high density of fish within these environments disturbs the acoustic measurements \cite{rundtop2016experimental}. Recent research has explored the use of stereo vision systems and image processing techniques to enhance UUV localization \cite{Skaldebo:2023:ROV}. Stereo cameras, for example, have been employed to achieve 3D spatial awareness, which is crucial in environments where precise positioning relative to net structures is required. Techniques such as the FFT method for relative pose estimation in net pens \cite{schellewald2021vision} and the TRU-depth network for depth estimation \cite{10611007} have shown promising results in underwater applications. Additionally, methods for pose estimation using laser triangulation have demonstrated accuracy comparable to DVL systems at a fraction of the cost, making them a viable option for short-distance ranging in fish farming environments. However, these methods also encounter similar issues with interference from fish and are additionally sensitive to light changes \cite{10185747_laser_triangulation}. 
 
To address these challenges, this paper investigates vision-based methods to: 1) obtain the relative 3D pose of an UUV from a flexible and deformable structure to facilitate control strategies for autonomous net inspection operations in fish farms, 2) construct the depth map from mono-vision data that can be crucial for both collision-free autonomous operations with UUVs in dynamic environments and for obtaining the real-time map of the inspected area to identify irregularities such as holes, or biofouling in net pens, 3) estimate the global pose of an UUV within the net pen utilizing the available sensor measurements and relative poses of the robot, and 4) create a detailed 3D map of the net-pen environment utilizing data obtained from an industrial scale fish farms.

The preliminary results showcase the potential of combining TRU-depth  \cite{10611007}, FFT \cite{schellewald2021vision}, and Wavemap \cite{reijgwart2023efficientvolumetricmappingmultiscale} to estimate both net-relative and global position of the UUV, and produce accurate maps, even in such dynamic environments.

\begin{figure*}[h]
  \centering
  \includegraphics[width=1\textwidth]{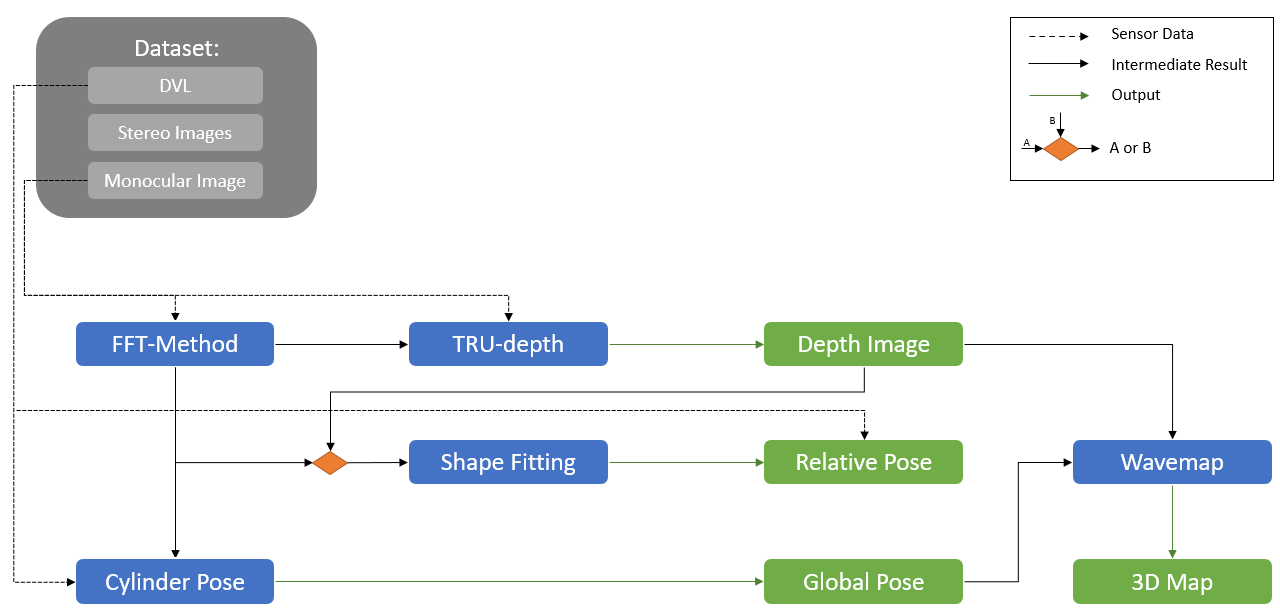}
  \caption{Overview of the general framework for localization and mapping for UUVs operating in dynamically changing environments}
  \label{fig:framework}
\end{figure*}
\section{Proposed Framework}
\label{sec:prob}
This section presents the full framework and discusses each of the pipeline's components in detail. In particular, it covers how the robot's net-relative pose, depth images, and global pose are estimated and how they are combined to create a 3D map of the net pen's inspected area (Fig. \ref{fig:framework}).
\subsection{Field Trials and Datasets}
\begin{figure}
    \centering
    \includegraphics[width=0.46\textwidth]{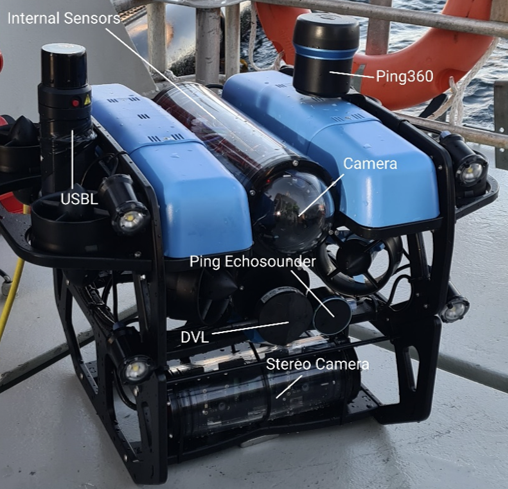}
    \caption{ROV with integrated sensors used in the field trials.}
    \label{fig:rov}
\end{figure}

Field trials have been performed in Rataren Cage 2 at SINTEF's ACE facilities \cite{sintef-ace-website}. The BlueROV2 with integrated sensors (Ping Echosounder, Ping360, Waterlinked DVL, Stereo Camera, Mono Camera), as shown in Fig. \ref{fig:rov},  has been deployed and was commanded to perform net-relative autonomous navigation using DVL measurements (see more information in \cite{haugalokken2024uuv}). During the trials, the net pen contained approximately 190.000 Atlantic Salmon. Several datasets have been recorded in different locations inside the net pen. In this paper, preliminary results are presented for one of the cases, where the vehicle was commanded to perform net-relative navigation at a distance of 1m, 2.1m, and 1.5m from the net, respectively.
\subsection{Net-relative Pose Estimation}
The first component of the proposed framework aims to estimate the net-relative position of the UUVs inside the net pen. The FFT method presented in \cite{schellewald2021vision} enables the net-relative pose estimation of an UUV, utilizing only monocular vision information. In particular, the FFT method analyzes the frequency spectrum of captured images to determine the distance and orientation of the camera based on characteristic regular patterns within the image, as well as the knowledge of the actual dimensions of the net squares. For a more detailed discussion on the original method, see \cite{schellewald2021vision}.

As shown in Fig. \ref{fig:framework}, the FFT method proposed in \cite{schellewald2021vision} has been modified in this paper to obtain multiple distance estimates to nets with a known grid cell size of $20 \times 20$ mm. Besides using the estimated distances from the modified FFT method as priors for TRU-depth, as described in the following section, the obtained 3D points have also been utilized to estimate the robot’s relative pose. In particular, instead of outputting a single position estimate, the modified version outputs multiple distance estimates, at known pixel locations. The outcome of the modified FFT provided the priors (distances) and then by applying plane and parabolic fitting and using the camera calibration parameters, it is possible to estimate the relative pose of the UUV.
\begin{figure*}
    \centering
    \includegraphics[width=1\textwidth]{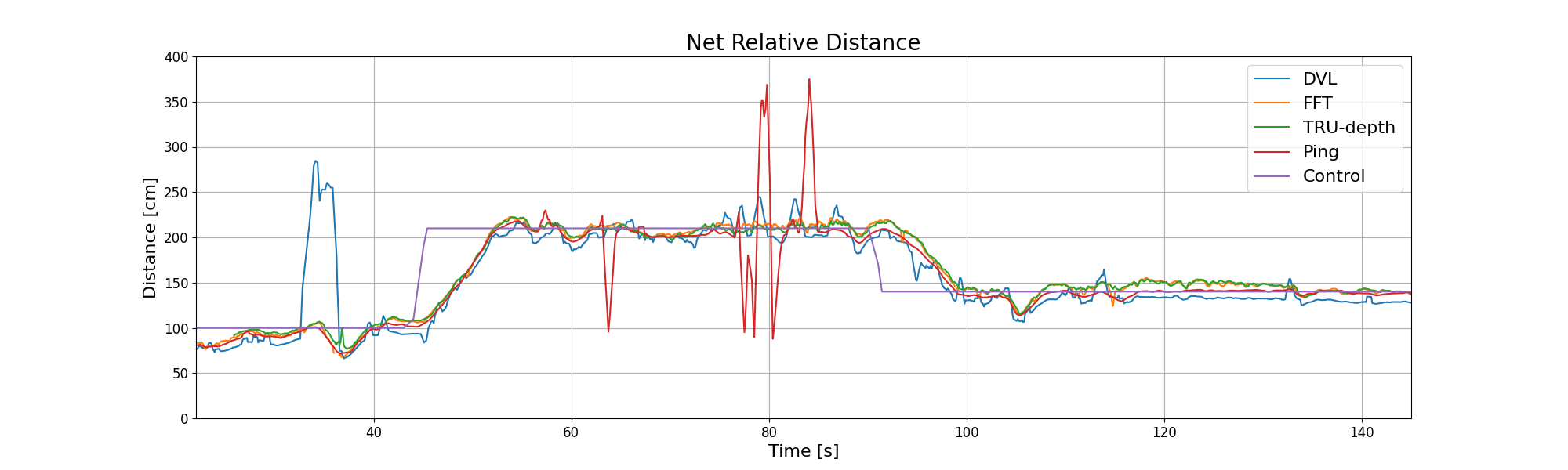}
    \caption{Net relative distance results.}
    \label{fig:distances}
\end{figure*}

\begin{figure}
    \centering
    \includegraphics[width=0.5\textwidth]{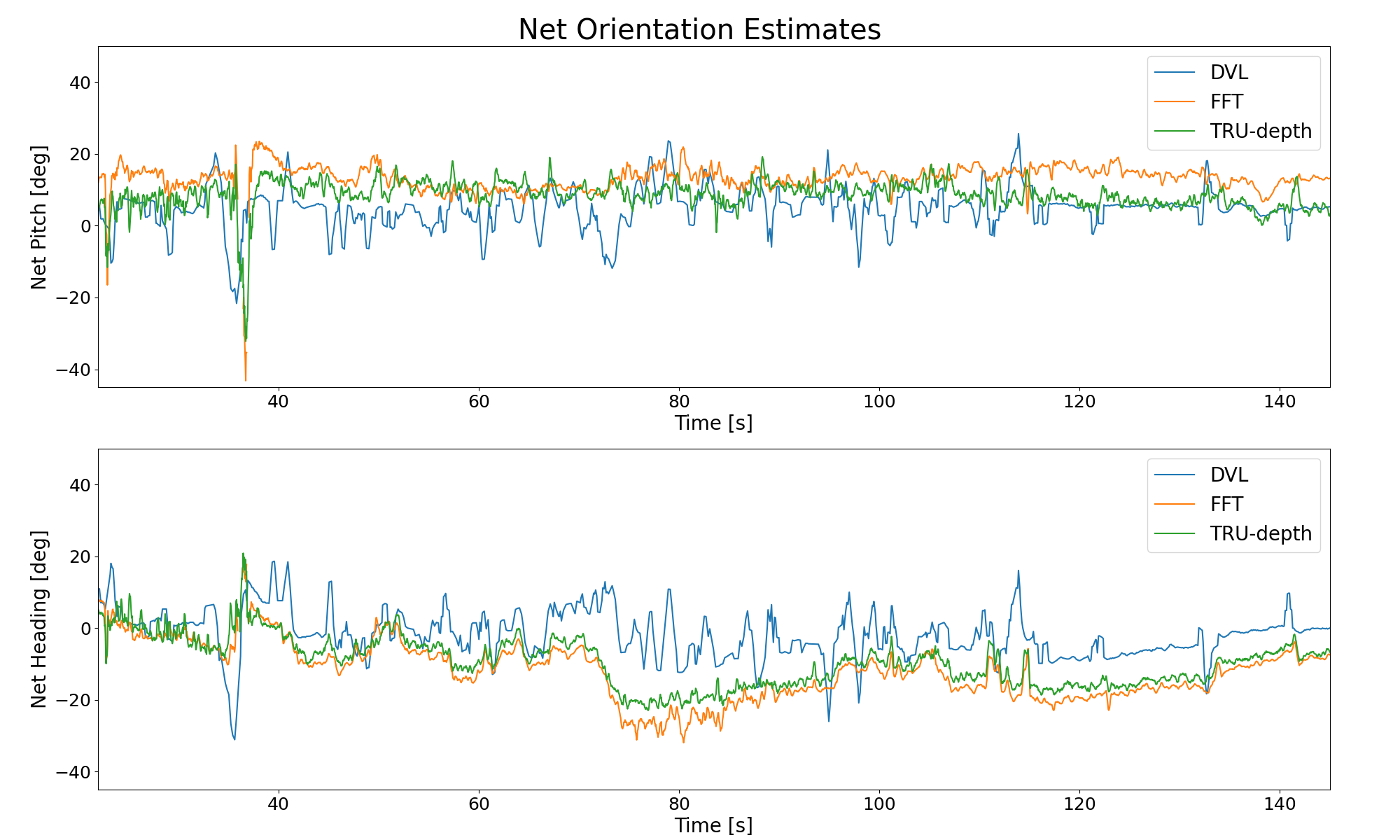}
    \caption{Net relative orientation results.}
    \label{fig:relative angles}
\end{figure}
\subsection{Depth Images}
In addition to focusing on net-relative poses, there is also interest in obtaining dense depth information, particularly for scene reconstruction and mapping purposes. Several methods have been demonstrated to be effective in other underwater environments \cite{10611007,10566210autonomous_uuvs}. This paper will discuss in detail the integration of the TRU-Depth method into the underwater localization and mapping framework.

The TRU-Depth method, proposed in \cite{10611007}, is a deep learning-based approach to generate dense depth images from monocular RGB images and some additional sparse depth information. The network utilizes these sparse depth priors to mitigate scale ambiguity, which results in a metrically scaled depth image. For structures such as nets with repetitive grids, it is quite challenging to obtain the priors required by TRU-Depth using classic feature-matching methods. Therefore, in order to obtain reliable priors for the net, the modified FFT method has been employed. This method provides reliable, uniformly distributed, and accurate depth estimates of the net, which are then used as priors for the TRU-Depth network (See Fig. \ref{fig:framework}. With these priors, the TRU-Depth method was able to generate dense depth images that accurately represent the 3D shape of the net pen. In an attempt to further enhance the results, the network was re-trained on images from the newly obtained dataset. Due to the lack of absolute ground truth, DVL
net-relative distance measurements were used to create uni-colored depth maps as ground truth for the re-training.

Note that since the FFT method specifically detects only the net, the TRU-Depth network also focuses solely on the net, disregarding any fish or other objects that might partially occlude the view. It would be interesting for further research to explore how the system responds when a larger structure occludes a significant portion of the net, preventing the FFT method from detecting it in those areas. This scenario could provide insights into how TRU-Depth handles regions lacking priors but resembling the environment in its training data more closely.
\subsection{Global Pose Estimation}
As mentioned previously, the modified FFT method provides robust relative pose estimates and therefore, these data have also been utilized to estimate the global pose of the UUV. Initially, the obtained FFT points were fit to a cylinder under the assumption that the net pen exhibits no deformation. Note that this is a reasonable assumption for small deformations, and the results demonstrate the efficacy of the simple method for global pose estimation proposed in this paper. Initially, the global radial coordinate is estimated by fitting a circle to the FFT points projected onto the $xy$-plane in camera coordinates. Subsequently, the global position and relative heading are computed by transforming to the global frame. The $z$-coordinate is directly obtained from the pressure sensor readout. Since the roll and pitch of the robot were controlled to be zero during the trials, the problem can be reduced to 2D, reducing its complexity. Finally, the global heading angle is calculated by summing the angular coordinate and the relative heading of the robot. The angular coordinate of the UUV is obtained utilizing the DVL velocities since these measurements were much less noisy than the IMU sensor data of the BlueROV2. In particular,  integrating the velocities over one time-step results in a new position estimate in the cylindrical frame. Afterwards, the new radius is replaced with the computed radius for the current time-step, effectively rotating the situation according to the velocity integration.
\subsection{3D Map Representation}
Mapping is crucial for UUVs operating in fish farms, particularly for inspection purposes. By generating detailed maps of the underwater environment, UUVs can accurately document the spatial relationships between structures, assess the integrity of the net, and monitor the net's conditions over time. Real-time mapping can also greatly benefit mission planning for autonomous vehicles. 

Two mapping approaches were tested: first, stacking of RGB point clouds generated from global pose estimates and camera images, which allowed for visual inspection of stacking quality and, consequently, the precision of the pose estimates. Second, the Wavemap method was applied (see \cite{reijgwart2023efficientvolumetricmappingmultiscale}) to evaluate whether this technique, in combination with TRU-Depth depth images, provides a valuable mapping solution for underwater applications. Both methods have demonstrated real-time capability \cite{10611007,reijgwart2023efficientvolumetricmappingmultiscale}, suggesting they offer promising directions for future development.
\section{Results}
This section presents the results and analysis across four key areas: relative poses, depth image results, global pose estimates, and mapping. To improve clarity and reduce noise, a sliding window smoothing filter has been applied to the plots of relative distance and orientation measurements. This technique enhances trend visibility and data variation by averaging points within a defined sliding window, offering a more accurate representation of the methods' performance. In contrast, the results for depth images, global pose estimates, and mapping are presented without smoothing to maintain the integrity of the raw data.
\subsection{Net-relative Pose Estimation}
The distance measurements from the DVL and the forward-facing ping echo sounder are compared with distance estimates obtained from the modified FFT method and the TRU-Depth, which utilized FFT-generated priors (Figure \ref{fig:distances}). Since the DVL measurements are used as a reference signal by the UUV's controller, the vision-based methods are compared to these values due to the lack of additional ground truth data. The results generally show a close alignment among the methods, with particularly strong agreement between the TRU-Depth and FFT results.  As expected due to the presence of fish, it is notable that both acoustic sensors exhibit clear measurement outliers. This indicates the advantage of utilizing vision-based methods for robust localization in challenging underwater environments such the ones faced in fish farms.

In \ref{fig:relative angles}, the comparison between the measured relative orientation computed from the DVL beams \cite{Amundsen2022} and the relative orientation estimates calculated from the FFT points, as well as the TRU-Depth-generated depth maps, is presented. As shown, the overall trends are consistent, although the differences in relative orientation are larger than those observed in the relative distance estimates discussed above. It is also evident that all methods for obtaining relative orientations exhibit a significant amount of noise, with the acoustic sensor showing the most. Note that due to the lack of accurate ground truth data, a definitive assessment of which method provides a more precise estimation of the net's relative orientation is not possible. Generally, the increased noise or variability in the results could partially be attributed to limited tuning of the heading controller during the trials. Future improvements could include better tuning of the controller to enhance performance.

Overall, the net-relative pose estimation results presented in this paper highlight the error-proneness of acoustic sensors in fish farming environments, as well as the capability of vision-based systems when operating close to or interacting with net structures. This underscores the importance of investigating vision-based methods for operations in fish farms, which often require net-relative control strategies.
\subsection{Depth Images}
\begin{figure}
    \centering
    \includegraphics[width=0.48\textwidth]{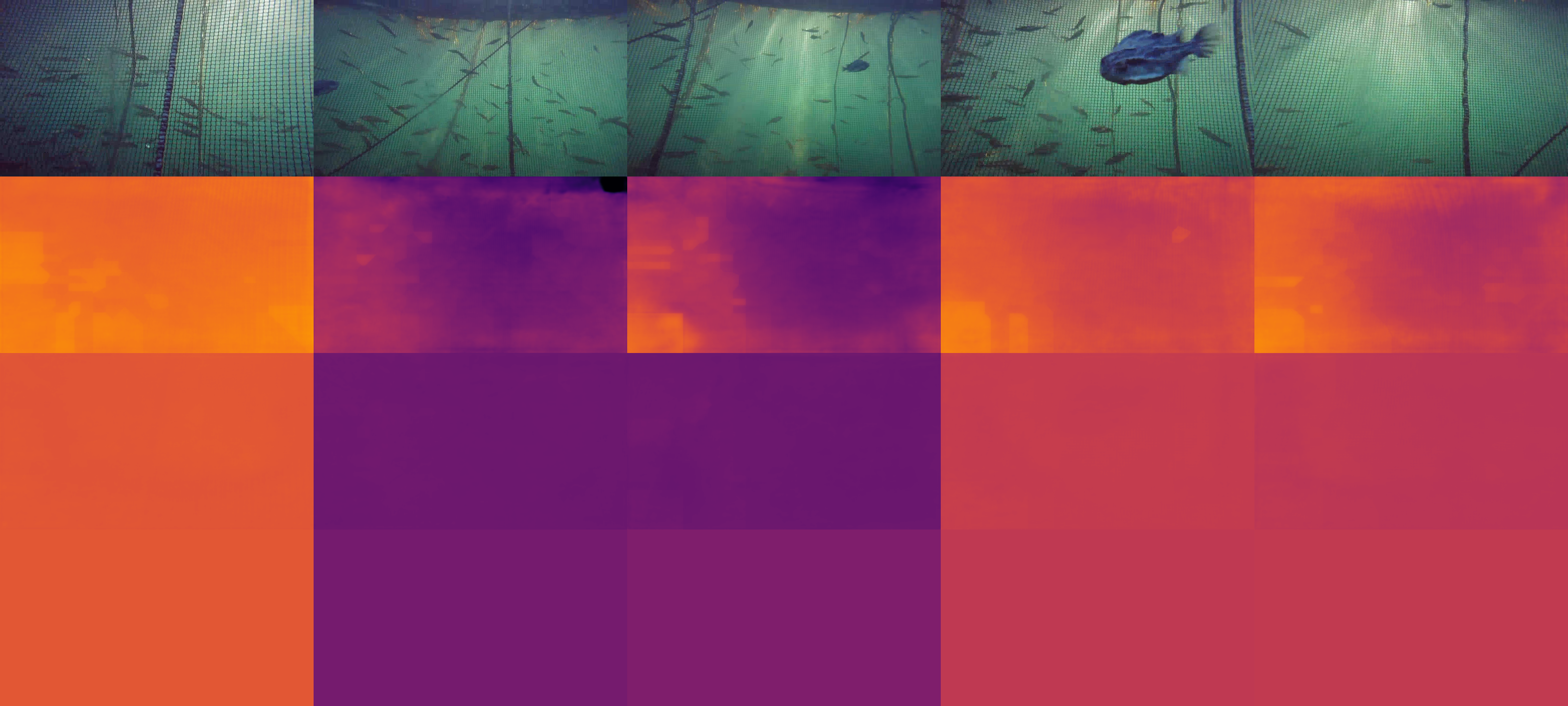}
    \caption{Comparison of depth images. 1: RGB Image; 2: TRU-Depth; 3: Retrained TRU-Depth; 4: DVL Reference.}
    \label{fig:depth image comparison}
\end{figure}

\begin{figure}
    \centering
    \includegraphics[width=0.48\textwidth]{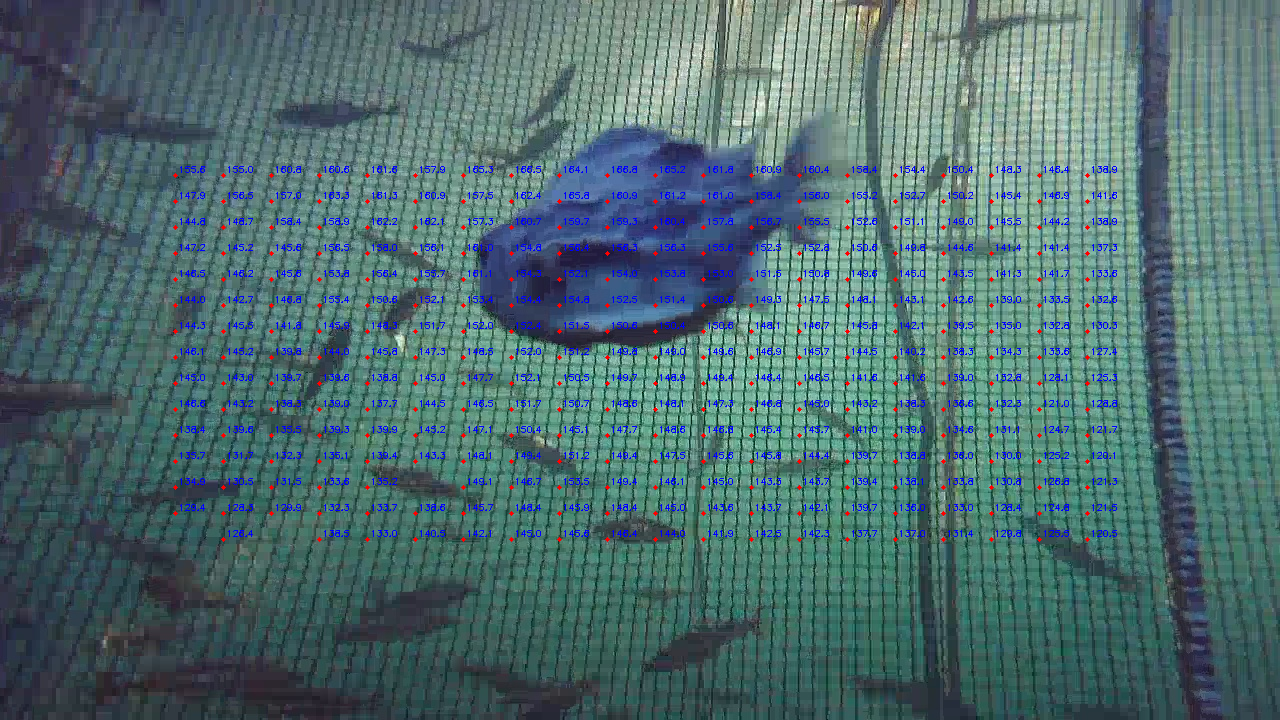}
    \caption{FFT generated priors under partially occluded conditions. (Values in [cm])}
    \label{fig:fft priors}
\end{figure}

This section compares the depth mapping results of TRU-Depth and the retrained TRU-Depth model. Though inspecting the resulting depth images visually (see Fig. \ref{fig:depth image comparison}), it is still notable that TRU-Depth network does not detect fish in front of the net, which was expected since the network is using FFT priors from the net (see Fig. \ref{fig:fft priors}). It is also visible (Fig. \ref{fig:depth image comparison}) that the retrained TRU-Depth model produces depth maps with a flatter appearance, almost uniform in color. Both versions of TRU-Depth yield depth images that are relatively close in color tone to the DVL measurements shown in the last row of Fig. \ref{fig:depth image comparison}. Note that the retrained TRU-Depth network, trained using the flat DVL depth maps as ground truth, shows a pronounced flattening effect after just one training epoch. While this flattening is expected due to the use of single-value depth maps, the extent of the effect is quite strong. The retraining does not significantly affect the overall distance estimates to the net but essentially removes the capability to estimate the net's relative orientation. Therefore, the original TRU-depth network from \cite{10611007} is applied in the proposed framework. The obtained results showcase the TRU-Depth network's effectiveness when provided with accurate priors, even in environments with few distinctive features and transparent structures not seen in its training data. This underscores the method's significance for fish farming operations, highlighting its potential to perform reliably in such challenging conditions.
\subsection{Global Pose Estimation}
\begin{figure}
    \centering
    {\includegraphics[width=0.48\textwidth]{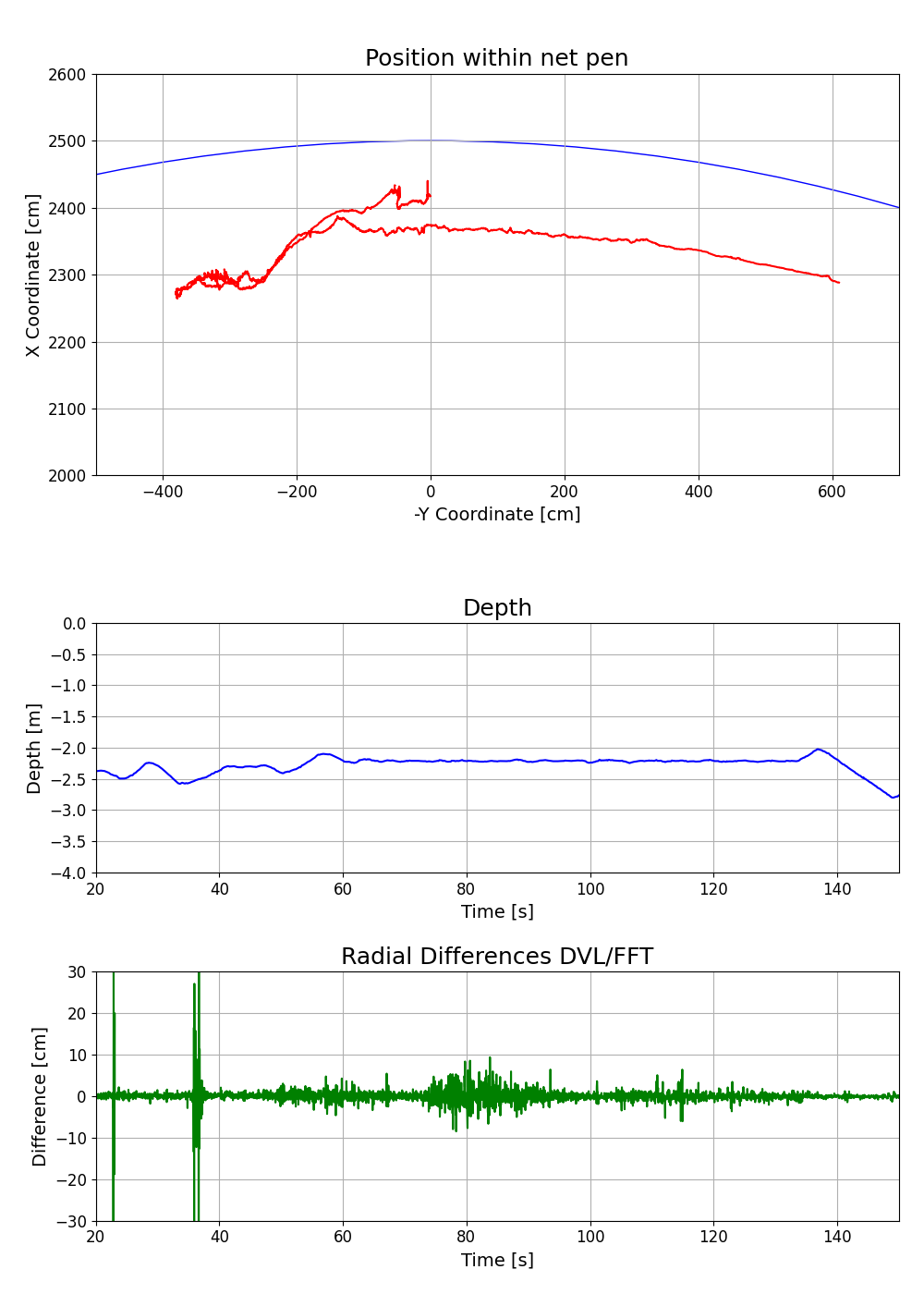}}
    \caption{Trajectory estimation results for the UUV.}
    \label{fig:trajectories}
\end{figure}

Assuming no deformations in the net pen and no pitch and roll of the UUV, the global position of the vehicle has been reconstructed using the DVL velocity measurements. The resulting trajectories are displayed in Fig. \ref{fig:trajectories}. The top plot shows the trajectory from a top-down view, while the second plot illustrates the third dimension, depth. The third plot displays the difference between the radial coordinate obtained from integration, and the one derived from circle fitting. It is evident that, aside from brief periods, the errors between the integration estimate and the actual optical distance estimate are small. The few peaks in the error correspond to instances where the ROV changes direction rapidly due to control input adjustments, resulting in imprecise DVL measurements and blurred images that lead to less accurate FFT estimations.

\begin{figure}
    \centering
    {\includegraphics[width=0.48\textwidth]{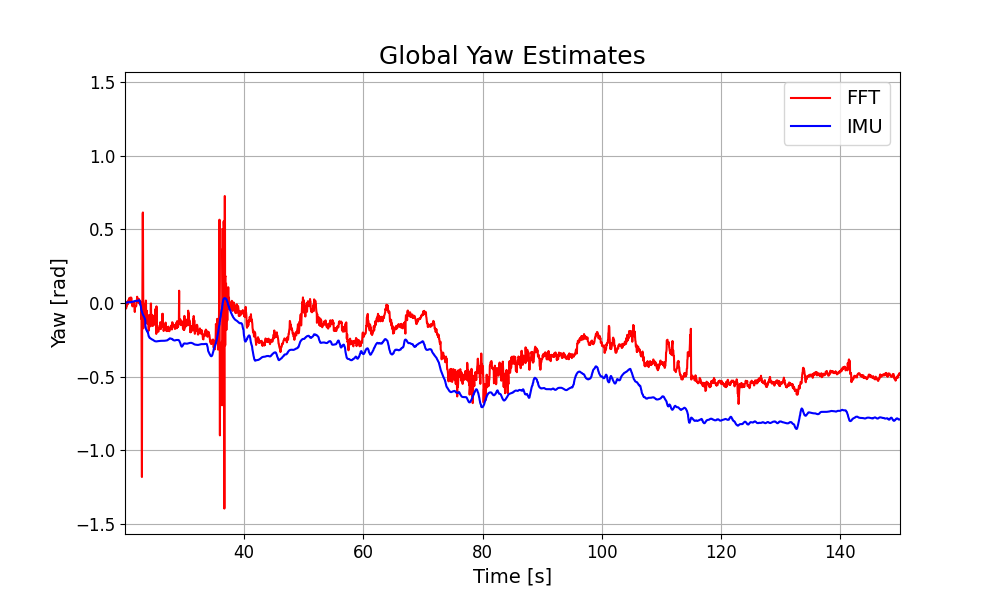}}
    \caption{Global Heading comparison of the IMU measurements and the cylinder pose estimations.}
    \label{fig:global yaw}
\end{figure}

To further evaluate the accuracy of the estimates, the calculated global yaw estimates have been compared with the onboard IMU measurements, as illustrated in Fig. \ref{fig:global yaw}. The plots are shifted so that the initial yaw value aligns, as the robot’s initial yaw was arbitrarily set to zero. The observed drift between the estimated and measured values over time could be attributed to several factors, including imprecise estimates, integration errors, or sensor drift in the DVL. Additionally, sensor drift in the IMU itself could contribute to the discrepancy. Overall, the vehicle's trajectory has been estimated in a manner that appeared consistent with the video data. Further evaluation through point cloud stacking indicated that the estimates are relatively accurate.

\begin{figure*}
    \centering
    {\includegraphics[width=1\textwidth]{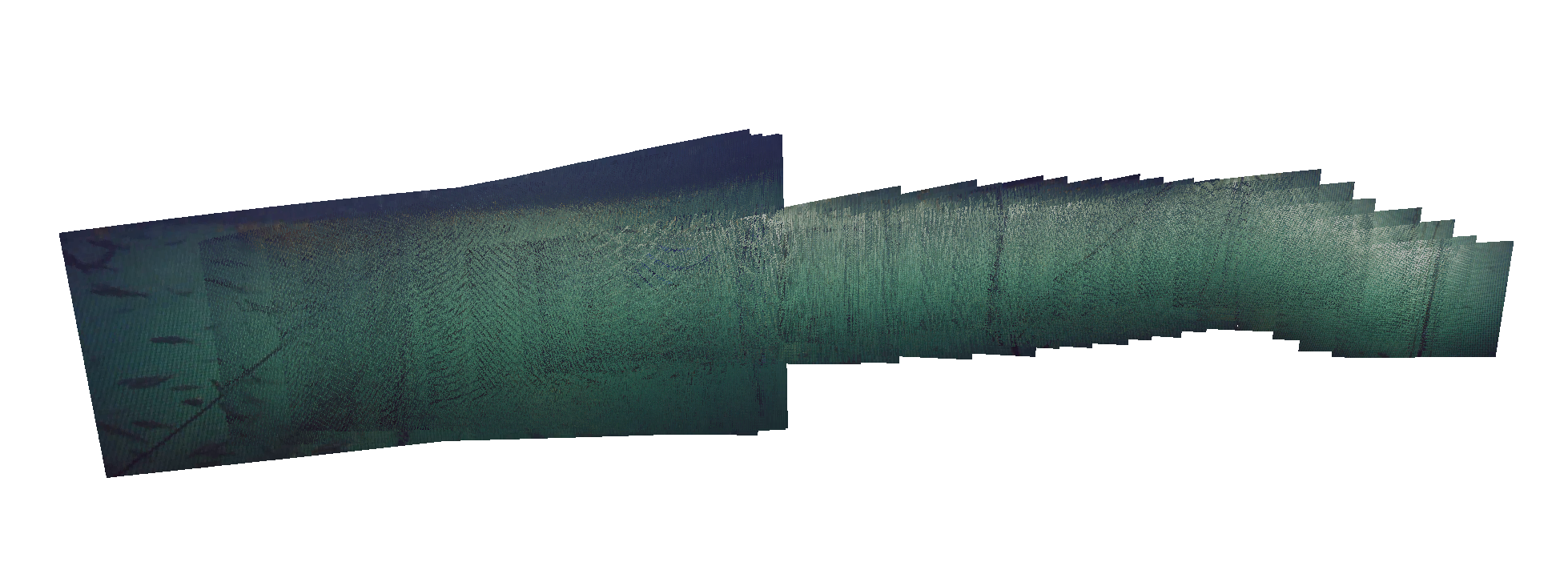}}
    \caption{Stacked point clouds from image projection onto cylinder fittings.}
    \label{fig:pointcloud stacks}
\end{figure*}
\begin{figure*}
    \centering
   {\includegraphics[width=1\textwidth]{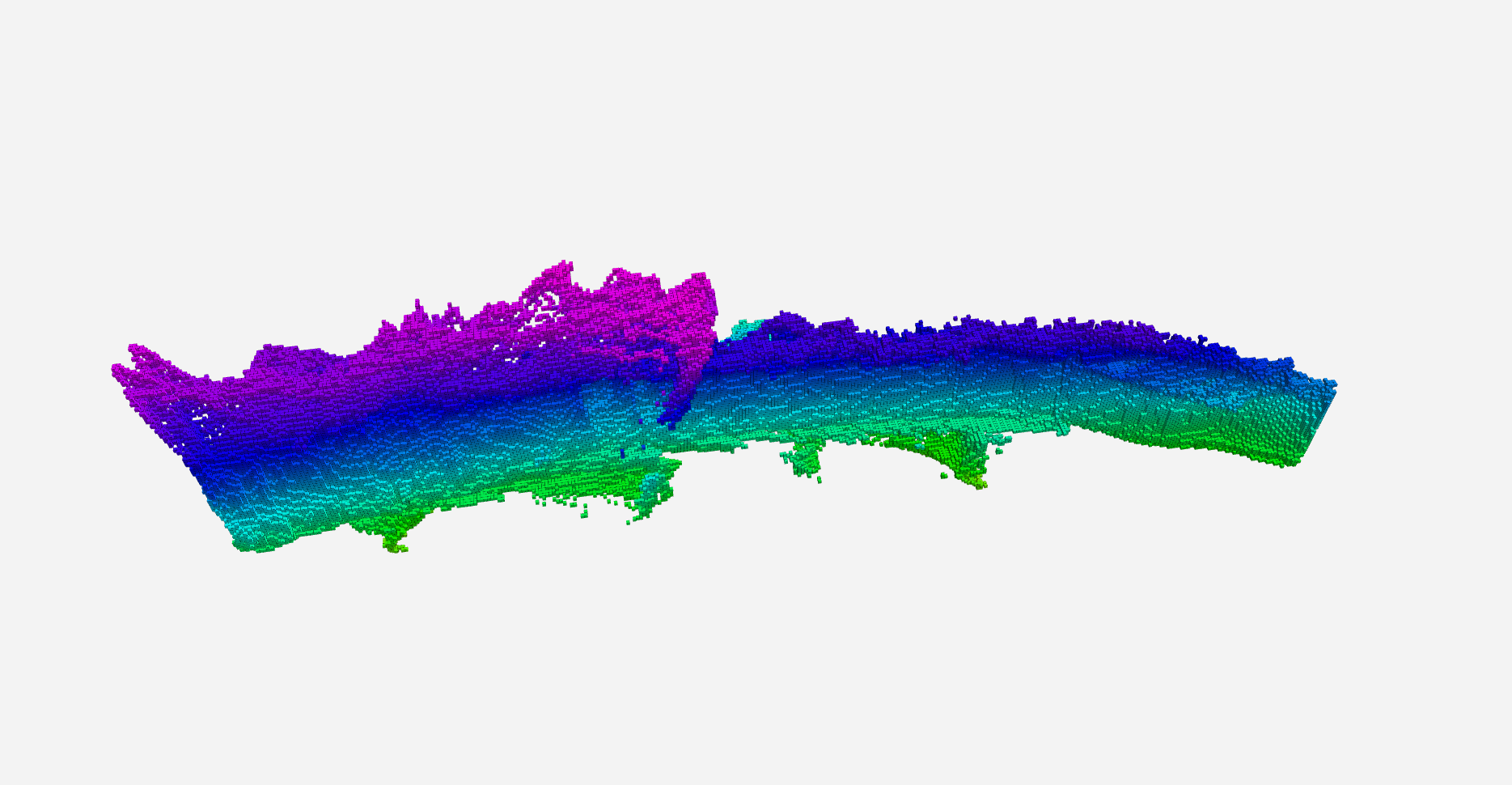}}
    \caption{Volumetric maps generated by the Wavemap method using TRU-Depth generated depth images and estimated 3D poses.}
    \label{fig:wavemap results}
\end{figure*}
\subsection{3D Map Representation}
To further assess the precision of the global pose estimations, the RGB data was projected onto the estimated net cylinder, and the resulting RGB point clouds were stacked, as shown in Fig. \ref{fig:pointcloud stacks}. By observing different lines visible across multiple images, one can gauge the accuracy of the position estimates. A clear example is the diagonal rope visible on the left side of the stacked point cloud (see Fig. \ref{fig:pointcloud stacks}), which runs through multiple images and connects smoothly, even though the images were taken from different distances to the net.

To assess the potential of using TRU-Depth-generated depth images for mapping and to evaluate the feasibility of applying the Wavemap method \cite{ reijgwart2023efficientvolumetricmappingmultiscale} in underwater environments, the Wavemap technique was employed on the TRU-Depth depth maps along with the estimated 3D poses. This resulted in the 3D volumetric maps visualized in Fig. \ref{fig:wavemap results}. The color gradient in this figure represents the value of the $z$ coordinate. The maps effectively display the net pen with minimal disturbances in the overlapping portions while distinct areas of increased noise are observed in the map.  The increased noise corresponds to images captured during the dive phase of the ROV, where all estimates, including direct DVL measurements, are extremely unreliable and noisy. Excluding these noisy sections, where the vehicle was not moving smoothly along the net, would directly improve map quality. In general, the results demonstrate the potential of combining the TRU-Depth method with the Wavemap method to create 3D maps of complex underwater environments for real-time applications, as both methods have been shown to operate in real-time \cite{10611007, reijgwart2023efficientvolumetricmappingmultiscale}.
\section{Conclusions}
\label{sec:conc}
This paper proposes a general, vision-based framework for underwater localization and mapping. The framework leverages the FFT to generate priors for TRU-Depth, which in turn provides the dense depth information required for 3D mapping. Additionally, methods for obtaining net-relative and global pose estimates of UUVs have been proposed. The preliminary results demonstrated the potential of the framework to integrate the FFT, TRU-Depth and Wavemap methods for applications in fish farming environments. It specifically showed that TRU-Depth can generate dense depth images from monocular images within these environments. When coupled with the Wavemap method and 3D pose estimates, the pipeline enables the creation of detailed volumetric maps. The completeness and accuracy of these 3D maps highlights their potential for real-world applications in the underwater domain. In the future, alternative methods to obtain priors of fish or other distinct structures could be combined with the FFT priors to enable TRU-Depth and Wavemap to comprehensively reconstruct 3D scenes from monocular images. 

\section{Acknowledgement}
The authors would like to thank Bent Haugaløkken, Sveinung J. Ohrem, Kay Arne Skarpnes, Terje Bremvåg and Linn Evjemo for their support on obtaining the data utilized in this paper.


\bibliographystyle{IEEEtran}
\bibliography{./IEEEabrv,refs}


\end{document}